\documentclass[11pt,a4paper]{article}
\usepackage[hyperref]{emnlp2018}
\usepackage{times}
\usepackage{latexsym}
\usepackage{url}
\usepackage{booktabs} 
\usepackage{float}
\usepackage{textcomp}
\usepackage{latexsym}
\usepackage{graphicx} 
\usepackage{microtype}
\usepackage{pifont}
\usepackage[font=small, labelfont=bf]{caption}
\usepackage[]{subcaption}
\usepackage{verbatim}
\usepackage{framed}
\usepackage{color, colortbl}
\usepackage[colorinlistoftodos]{todonotes}
\usepackage{amssymb}
\usepackage{amsmath}
\usepackage{capt-of}
\usepackage{multirow}
\usepackage{algorithmicx}
\usepackage{algorithm}
\usepackage{algcompatible}
\usepackage{paralist}
\usepackage{enumitem}
\usepackage{balance}
\usepackage{tabularx}
\usepackage{bm}
\usepackage{lipsum}
\usepackage[all]{nowidow}

\definecolor{high}{HTML}{FFCE8E}

\DeclareMathOperator{\E}{\mathbb{E}}

\newcommand\blfootnote[1]{%
	\begingroup
	\renewcommand\thefootnote{}\footnote{#1}%
	\addtocounter{footnote}{-1}%
	\endgroup
}

\aclfinalcopy 

\title{Multi-view Models for Political Ideology Detection of News Articles}

\author{Vivek Kulkarni \\
    Department of Computer Science \\ 
  	University of California, Santa Barbara \\
  {\tt vvkulkarni@cs.ucsb.edu} \\ \And
  Junting Ye \\
   Department of Computer Science \\ 
   Stony Brook University \\
  {\tt juyye@cs.stonybrook.edu} \\ \AND
  Steven Skiena \\
   Department of Computer Science \\ 
   Stony Brook University \\
  {\tt skiena@cs.stonybrook.edu} \\ \And
  William Yang Wang \\
    Department of Computer Science \\ 
  	University of California, Santa Barbara \\
  {\tt william@cs.ucsb.edu} \\
}

\date{}

\begin{document}
\maketitle
\begin{abstract}
A news article's title, content and link structure often reveal its political ideology. However, most existing works on automatic political ideology detection only leverage textual cues. Drawing inspiration from recent advances in neural inference, we propose a novel attention based multi-view model to leverage cues from all of the above views to identify the ideology evinced by a news article. Our model draws on advances in representation learning in natural language processing and network science to capture cues from both textual content and the network structure of news articles. We empirically evaluate our model against a battery of baselines and show that our model outperforms state of the art by $10$ percentage points F1 score. \blfootnote{\copyright\ The authors, 2018. This draft is the author’s draft of the paper and has been posted here for your personal use.}
\end{abstract}
\section{Introduction}
\label{sec:introduction}
Many issues covered or discussed by the media and politicians today are so subtle that even word-choice may require one to adopt a particular ideological position \cite{iyyer2014political}. For example, conservatives tend to use the term \texttt{tax reform}, while liberals use \texttt{tax simplification}. Though objectivity and unbiased reporting remains a cornerstone of professional journalism, several scholars argue that the media displays ideological bias \cite{gentzkow2010drives,groseclose2005measure,iyyer2014political}. Even if one were to argue that such bias may not be reflective of a lack of objectivity, prior research \citet{dardis2008media,card2015media} note that framing of topics can significantly influence policy.

\begin{figure*}[t!]
	\centering
	\begin{subfigure}{0.45\textwidth}
		\includegraphics[width=0.8\textwidth]{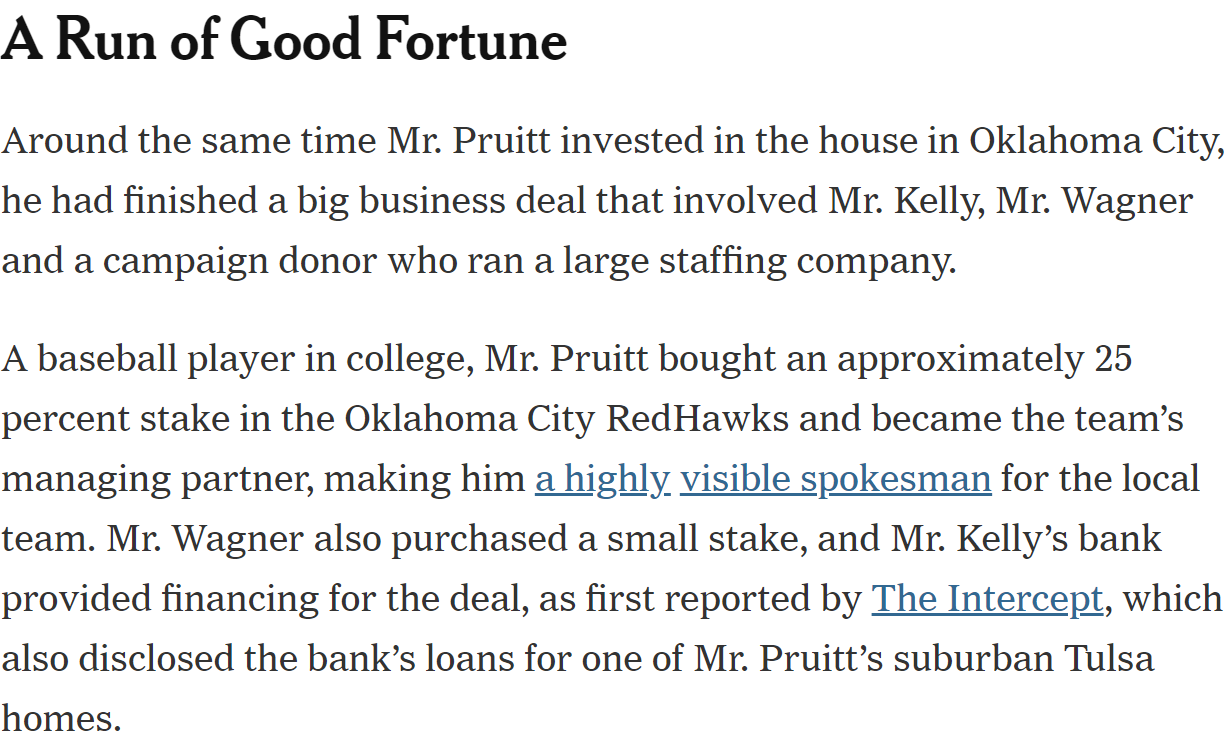}
		\caption{A sample news article. Note the presence of hyperlinks to other sources like The Intercept.}
		\label{fig:news_article}
	\end{subfigure}\hfill
	\begin{subfigure}{0.45\textwidth}
		\centering
		\includegraphics[width=0.5\textwidth]{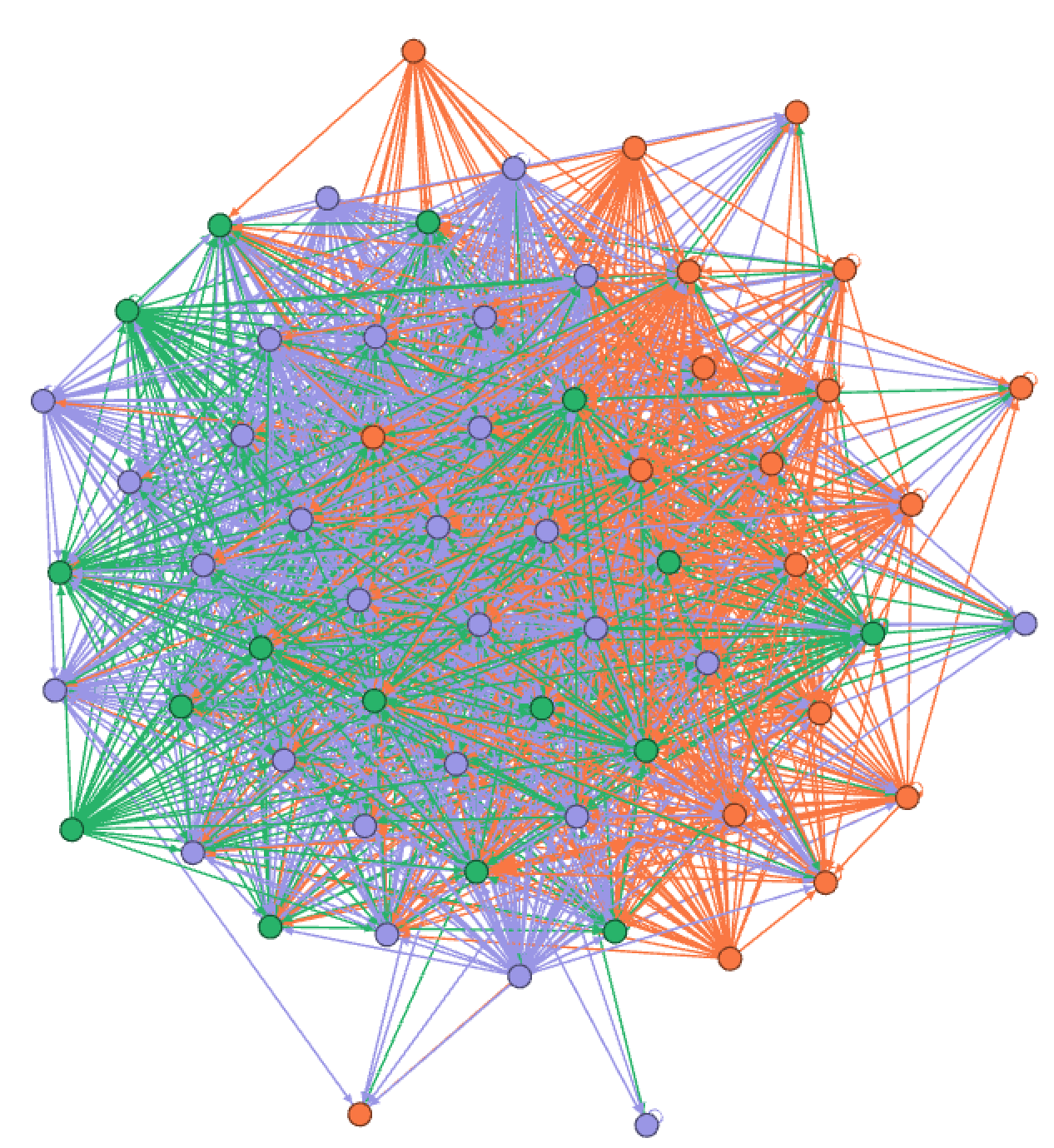}
		\caption{Homophily in link structure (viewed in color) of various news sources which can be observed by noting the presence of clusters corresponding to political ideologies. The blue, orange and green clusters correspond to left, right and center leaning sources respectively.}
		\label{fig:network}
	\end{subfigure}
	\caption{Our proposed framework MVDAM models multiple views of the news article including the content and the link structure. Figure \ref{fig:news_article} shows a sample article from the New York Times. The presence of such links can provide informative signals for predictive tasks like ideology detection primarily due to homophily (Figure \ref{fig:network}).}
	\label{fig:crown_jewel}
\end{figure*}

Since manual detection of political ideology is challenging at a large scale, there has been extensive work on developing computational models for automatically inferring the political ideology of articles, blogs, statements, and congressional speeches \cite{gentzkow2010drives,iyyer2014political,preoctiuc2017beyond,sim2013measuring}. In this paper, we consider the detection of ideological bias at the news article level, in contrast to recent work by \citet{iyyer2014political} who focus on the sentence level or the work of \cite{preoctiuc2017beyond} who focus on inferring ideological bias of social media users. Prior research exists on detecting ideological biases of news articles or documents \cite{gentzkow2010drives,gerrish2011predicting,iyyer2014political}. However, all of these works generally only model the text of the news article.  However, in the online world, news articles do not just contain text but have a rich structure to them. Such an online setting influences the article in subtle ways: (a) choice of the title since this is what is seen in snippet views online (b) links to other news media and sources in the article and (c) the actual textual content itself. Except for the textual content, prior models ignore the rest of these cues. Figure \ref{fig:crown_jewel} shows an example from \texttt{The New York Times}. Note the presence of hyperlinks in the text, which link to other sources like \texttt{The Intercept} (Figure \ref{fig:news_article}).  We hypothesize that such a link structure is reflective of homophily between news sources sharing similar political ideology -- homophily which can be exploited to build improved predictive models (see Figure \ref{fig:network}). Building on this insight, we propose a new model \textbf{MVDAM}: \emph{Multi-view document attention model} to detect the ideological bias of news articles by leveraging cues from multiple views: the title, the link structure, and the article content. Specifically, our contributions are:
\begin{enumerate}[noitemsep]
	\item We propose a generic framework \textbf{MVDAM} to incorporate multiple views of the news article and show that our model outperforms state of the art by $10$ percentage points on the F1 score.
	\item  We propose a method to estimate the ideological proportions of sources and rank them by the degree to which they lean towards a particular ideology.
	\item Finally, differing from most works, which typically focus on congressional speeches, we conduct ideology detection of news articles by assembling a \emph{large-scale diverse dataset} spanning more than $50$ sources. 
\end{enumerate}

\section{Related Work}
\label{sec:related_work}
Several works study the detection of political ideology through the lens of computational linguistics and natural language processing \cite{laver2003extracting,monroe2004talk,thomas2006get,lin2008joint,carroll2009measuring,ahmed2010staying,gentzkow2010drives,gerrish2011predicting,sim2013measuring,iyyer2014political,preoctiuc2017beyond}. \citet{gentzkow2010drives} first attempt to rate the ideological leaning of news sources by proposing a measure called ``slant index'' which captures the degree to which a particular newspaper uses partisan terms or co-allocations.  \citet{gerrish2011predicting} predict the
voting patterns of Congress members based on supervised topic models while \citet{ahmed2010staying,lin2008joint} use similar models to predict bias in news articles, blogs, and political speeches \cite{iyyer2014political}. Differing from the above, \citet{sim2013measuring} propose a novel HMM-based model to infer the ideological proportions of the rhetoric used by political candidates in their campaign speeches which relies on a fixed lexicon of bigrams.

The work that is most closely related to our work is that of \citet{iyyer2014political,preoctiuc2017beyond}. \citet{iyyer2014political} use recurrent neural networks to predict political ideology of congressional debates and articles in the ideological book corpus (IBC) and demonstrate the importance of compositionality in predicting ideology where modifier phrases and punctuality affect the political ideological position. \citet{preoctiuc2017beyond} propose models to infer political ideology of Twitter users based on their everyday language. Most crucially, they also show how to effectively use the relationship between user groups to improve prediction accuracy. Our work draws inspiration from both of these works but differentiates itself from these in the following aspects: We leverage the structure of a news article by noting that an article is just not free-form text, but has a rich structure to it. In particular, we model cues from the title, the inferred network, and the content in a joint generic neural variational inference framework to yield improved models for this task. Furthermore, differing from \citet{iyyer2014political}, we also incorporate attention mechanisms in our model which enables us to inspect which sentences (or words) have the most predictive power as captured by our model.  Finally, since we work with news articles (which also contain hyperlinks), naturally our setting is different from all other previous works in general (which mostly focus on congressional debates) and in particular from \citet{iyyer2014political} where only textual content is modeled or \citet{preoctiuc2017beyond} which focuses on social media users.
\section{Dataset Construction}
\label{sec:data}
\paragraph{News Sources} We rely on the data released by \textsc{AllSides.com}\footnote{https://www.allsides.com/media-bias/media-bias-ratings} to obtain a list of $59$ US-based news sources along with their political ideology ratings: \textsc{Left}, \textsc{Center} or \textsc{Right} which specify our target label space.  While we acknowledge that there is no ``perfect'' measure of political ideology, \textsc{Allsides.com} is an apt choice for two main reasons. First, and most importantly the ratings are based on a blind survey, where readers are asked to rate news content without knowing the identity of the news source or the author being rated. This is also precisely the setting in which our proposed computational models operate (where the models have access to the content but are agnostic of the source itself) thus seeking to mirror human judgment closely. Second, these are normalized by \textsc{AllSides} to ensure they closely reflect popular opinion and political diversity present in the United States. These ratings also correlate with independent measurements made by the \textsc{Pew Research Centre}. All these observations suggest that these ratings are fairly robust and generally ``reflective of the average judgment of the American People''\footnote{https://www.allsides.com/media-bias/about-bias}. 

\paragraph{Content Extraction}
Given the set of news sources selected above, we extract the article content for these news sources. We control for time by obtaining article content over a fixed time-period for all sources.  Specifically, we spider several news sources and perform data cleaning. In particular, the spidering component collates the raw HTML of news sources into a storage engine (MongoDB). We track thousands of US based news outlets including country wide popular news sources as well as many local/state news based outlets like the Boston Herald\footnote{This is a part of an ongoing project called MediaRank. More details can be found at \url{http://media-rank.com}}. However, in this paper, we consider only the $59$ US news sources for which we can derive ground truth labels for political ideology. For each of the news sources considered, we extract the title, the cleaned pre-processed content, and the hyperlinks within the article that reveal the network structure. The label for each article is the label assigned to its source as obtained from \textsc{AllSides}. We choose a random sample of $120,000$ articles and create 3 independent splits for training ($100,000$), validation ($10,000$) and test ($10,000$) with a roughly balanced label distribution. \footnote{Note that we do not restrict the articles to be strictly political since even articles on other topics like health and sports can be reflective of political ideology \cite{hoberman1977sport}.}  

\paragraph{Data Pre-processing and Cleaning} Since the labels were derived from the source, we are careful to remove any systematic features in each article which are trivially reflective of the source, since that would result in over-fitting. In particular we perform the following operations: (a) \textbf{Remove source link mentions} When modeling the link structure of an article, we explicitly remove any link to the source itself. Second, we also explicitly remove any systematic link structures in articles that are source specific. In particular, some sources may always have links to other domains (like their own franchisees or social media sites). These links are removed explicitly by noting their high frequency.
(b) \textbf{Remove headers, footers, advertisements} News sources systematically introduce footers, and advertisements which we remove explicitly. For example, every article of the \texttt{The Daily Beast} has the following footer \texttt{You can subscribe to the Daily Beast here} which we filter out.
\section{Models and Methods}
\paragraph{Problem Formulation} Given $\bm{X}=\{\bm{X_{title}},\bm{X}_{net},\bm{X_{content}}\}$ which represents a set of multi-modal features of news articles and a label set $\bm{Y}=\{\textsc{Left},\textsc{Center},\textsc{Right}\}$, we would like to model $\Pr(\bm{Y}|\bm{X})$. 

\begin{figure*}[t!]
\begin{subfigure}{0.5\textwidth}
\centering
	\includegraphics[width=\textwidth]{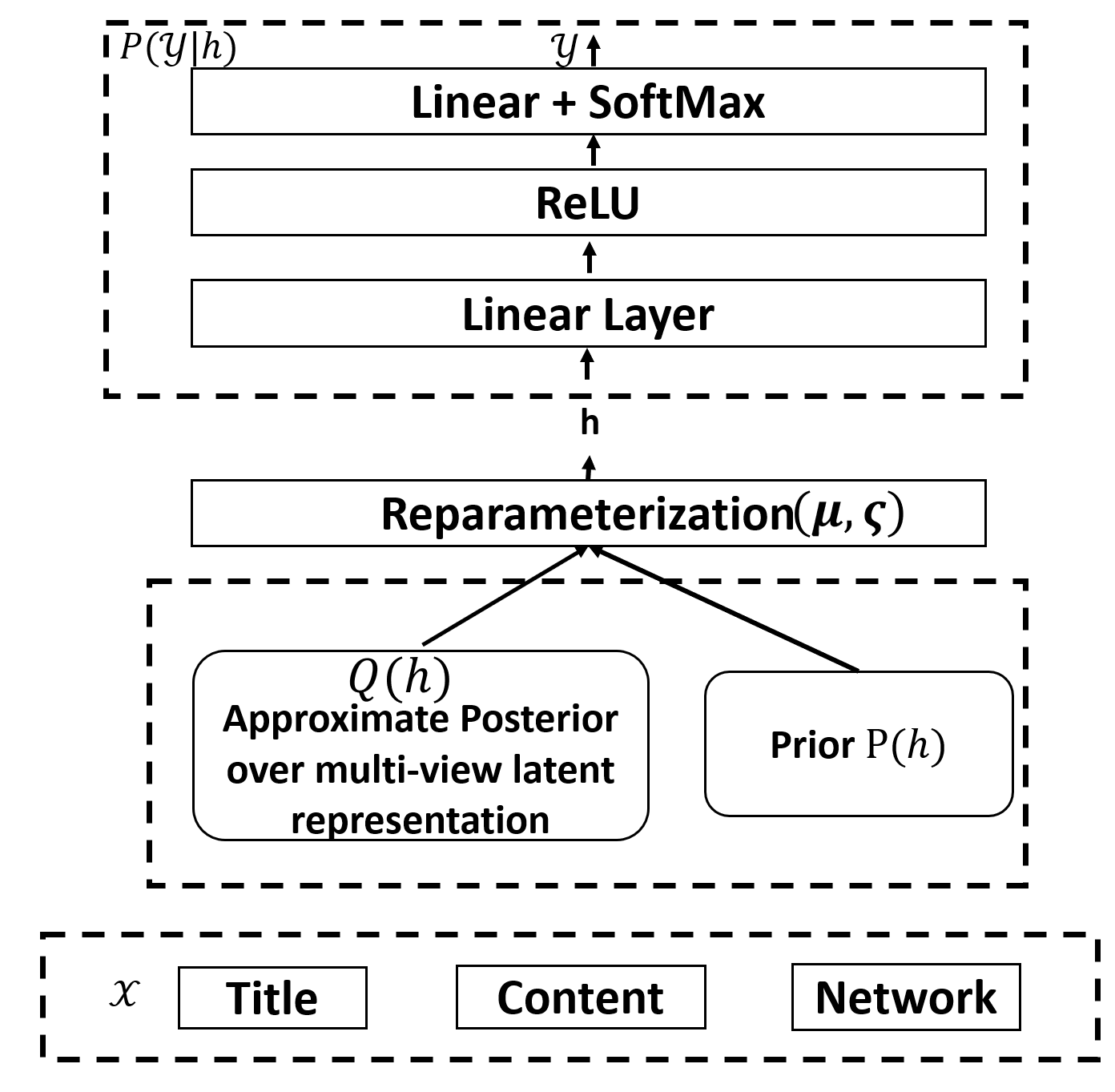}
	\caption{Overview of our full model.}
	\label{fig:our_model}
\end{subfigure}
\begin{subfigure}{0.5\textwidth}
\centering
	\includegraphics[width=\textwidth]{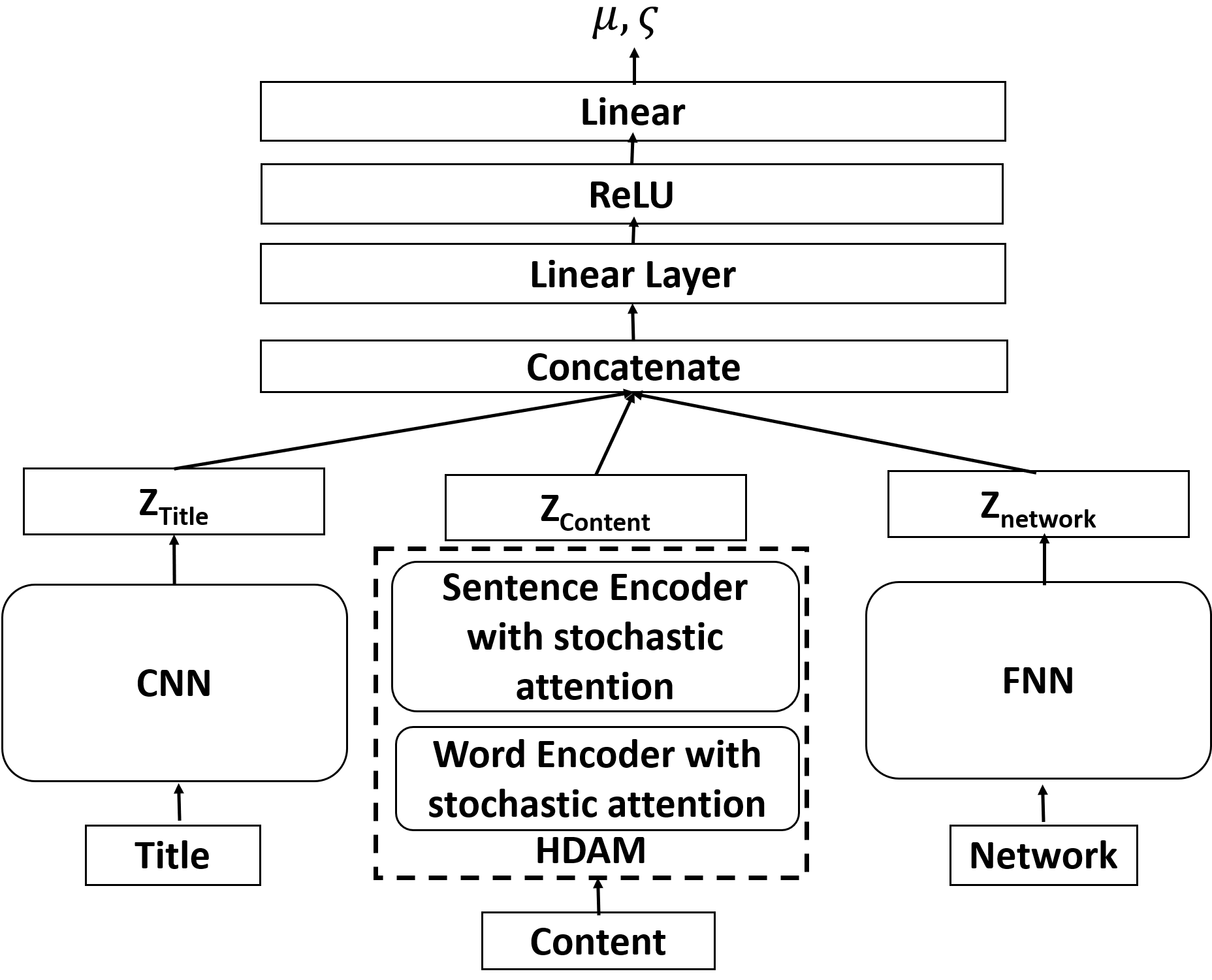}
	\caption{Overview of the inference network.}
	\label{fig:approx_posterior}
\end{subfigure}
\caption{A broad overview of our MVDAM model depicting the three major components:a discriminator, an inference network and a prior and captures cues from multiple views of the news article. As noted by \citet{miao2016neural} we use stochastic attention units which are shown to model ambiguity better. We thus train the model end-to-end using neural variational inference.}
\end{figure*}

\paragraph{Overview of MVDAM} We consider a Bayesian approach with stochastic attention units to effectively model textual cues. Bayesian approaches with stochastic attention have been noted to be quite effective at modeling ambiguity as well as avoiding over-fitting scenarios especially in the case of small training data sets \cite{miao2016neural}.
In particular, we assume a latent representation $\bm{h}$ learned from the multiple modalities in $\bm{X}$ which is then mapped to the label space $\bm{Y}$. In the most general setting, instead of learning a deterministic encoding $\bm{h}$ given $\bm{X}$, we posit a latent distribution over the hidden representation $\bm{h}$, $\Pr(\bm{h}|\bm{X})$ to model the overall document where $\Pr(\bm{h}|\bm{X})$ is parameterized by a diagonal Gaussian distribution $\mathcal{N}(\bm{h}|\mu(\bm{X}), \sigma^{2}(\bm{X}))$. 

Specifically, consider the distribution $\Pr(\bm{Y|X})$ which can be written as follows:
\begin{equation}
    \Pr(\bm{Y|X})=\sum_{\bm{h}}\Pr(\bm{Y}|\bm{h})\Pr(\bm{h}|\bm{X})
\end{equation}
As noted by \citet{miao2016neural}, computing the exact posterior is in general intractable. Therefore, we posit a variational distribution $q_{\phi}(\bm{h})$ and maximize the evidence lower bound $\mathcal{L} \leq \Pr(\bm{y}|\bm{X})$ namely, 
\begin{equation}
    \mathcal{L}=\E_{q_{\phi}(\bm{h})}[p(\bm{Y}|\bm{h})]-D_{KL}(q_{\phi}(\bm{h})||p(\bm{h|\bm{X}})),
    \label{eq:lvb}
\end{equation} where $p(\bm{Y}|\bm{h})$ denotes a probability distribution over $\bm{Y}$ given the latent representation $\bm{h}$, and $p(\bm{h}|\bm{X})$ denotes the probability distribution over $\bm{h}$ conditioned on $\bm{X}$.  

Equation \ref{eq:lvb} can be interpreted as consisting of three components, each of which can modeled separately: (a) \textbf{Discriminator} $p(\bm{Y}|\bm{h})$ can be viewed as a discriminator given the hidden representation $\bm{h}$. Maximizing the first term is thus equivalent to minimizing the cross-entropy loss between the model's prediction and true labels. (b) The second term, the KL Divergence term consists of two components: (1) \textbf{Approximate Posterior} The term $q_{\phi}(\bm{h})$ also known as the approximate posterior parameterizes the latent distribution which encodes the multi-modal features $\bm{X}$ of a document. (2) \textbf{Prior} The term $p(\bm{h}|\bm{X})$ can be viewed as a prior which can be uninformative (a standard Gaussian prior in the most general case, or any other prior model based on other features). We now discuss how we model each of these components in detail.

\subsection{Discriminator} We use a simple feed-forward network with a linear layer that accepts as input the latent hidden representation of $\bm{X}$, followed by a ReLU for non-linearity followed by a linear layer and a final soft-max layer to model this component.   

\subsection{Approximate Posterior} Here we model the approximate posterior $q_{\phi}(\bm{h})$ by an inference network shown succinctly in Figure \ref{fig:approx_posterior}. The inference network takes as input the features $\bm{X}$ and learns a corresponding hidden representation $\bm{h}$. More specifically, it outputs two components: ($\mu$, $\varsigma$) corresponding to the mean and log-variance of the gaussian parametrizing the hidden representation $\bm{h}$. We model this using a ``multi-view'' network which incorporates hidden representations learned from multiple modalities into a joint representation. Specifically, given $d$-dimensional hidden representations corresponding to multiple modalities $\bm{z}_{title}$, $\bm{z}_{network}$, and $\bm{z}_{content}$ the model first concatenates these representations into a single $3d$-dimensional representation $\bm{z}_{concat}$ which is then input through a 2-layer feed-forward network to output a $d$-dimensional mean vector $\mu$ and a $d$-dimensional log-variance vector $\varsigma$ that parameterizes the latent distribution governing $\bm{h}$. We now discuss the models used for capturing each view.

\subsubsection{Modeling the Title}
We learn a latent representation of the title of a article by using a convolutional network. Convolutional networks have been shown to be very effective for modeling short sentences like titles of news articles. In particular, we use the same architecture proposed by \cite{kim2014convolutional}.  The input words of the title are mapped to word embeddings and concatenated and passed through convolutional filters of varying window sizes. This is then followed by a max-over-time pooling \cite{collobert2011natural}. The outputs of this layer are input to a fully connected layer of dimension $d$ with drop-out which outputs $\bm{z}_{title}$, the latent representation of the title.
\subsubsection{Modeling the Network Structure of articles}
Capturing the network structure of article consists of two steps: (a) Learning a network representation of each source based on its social graph $G$. (b) Using the learned representation of each source to capture the link structure of a particular article. 

We use a state-of-the-art network representation learning algorithm to learn representations of nodes in a social network. In particular, we use Node2Vec \cite{node2vec-kdd2016}, which learns a $d$-dimensional representation of each source given the hyperlink structure graph $G$. 
Node2Vec seeks to maximize the log likelihood of observing the neighborhood of a node $\mathcal{N}(u)$, given the node $u$. Let $\bm{F}$ be a matrix of size $(V,d)$ where $\bm{F}(u)$ represents the embedding of node $u$. We then maximize the following likelihood function $\max_{\bm{F}} \sum_{u} \log \Pr(\mathcal{N}(u)|u)$.
We model the above likelihood similar to the Skip-gram architecture \cite{mikolov2013efficient} by assuming that the likelihood of observing a node $v\in \mathcal{N}(u)$ is conditionally independent of any other node in the neighborhood given $u$. That is $\log \Pr(\mathcal{N}(u)|u)=\sum_{v\in \mathcal{N}(u)}\log\ \Pr(v|u)$. We then model $\Pr(v|u)=\frac{e^{\bm{F}(u).\bm{F}(v)}}{\sum_{v}e^{\bm{F}(u).\bm{F}(v)}}$. Having fully specified the log likelihood function, we can now optimize it using stochastic gradient ascent. 

Having learned the embedding matrix $\bm{F}$ for each source node, we now model the link structure of any given article $\mathcal{A}$ simply by the average of the network embedding representations for each link $l$ referenced in $\mathcal{A}$. In particular, we compute $\bm{z}_{network}$ as: $\bm{z}_{network}=\frac{1}{|\mathcal{A}|}\sum_{l\in \mathcal{A}}\bm{F}(l)$.  

\subsubsection{Modeling the Content of articles}
To model the content of an article, we use a hierarchical approach with attention. In particular, we compute attention at both levels: (a) words and (b) sentences. We closely follow the approach by \cite{yang2016hierarchical} which learns a latent representation of a document $d$ using both word and sentence attention models. 

We model the article $\mathcal{A}$ hierarchically, by first representing each sentence $i$ with a hidden representation $s_{i}$. We model the fact that not all words contribute equally in the sentence through a word level attention mechanism. We then learn the representation of the article $\mathcal{A}$ by composing these individual sentence level representations with a sentence level attention mechanism.  

\paragraph{Learning sentence representations} We first map each word to its embedding matrix through a lookup embedding matrix $\bm{W}$. We then learn a hidden representation of the given sentence $h_{it}$ centered around word $w_{i}$ by embedding the sentence through a bi-directional GRU as described by \cite{bahdanau2014neural}. Since not all words contribute equally to the representation of the sentence, we introduce a word level attention mechanism which attempts to extract relevant words that contribute to the meaning of the sentence. Specifically we learn a word level attention matrix $\bm{W_{w}}$ as follows $
\alpha_{i}\propto\exp({\bm{W_{w}}h_{it}+ b_{w}}), 
s_{i} = \sum_{t}\alpha_{i}h_{it}$ where $s_{i}$ is the latent representation of the sentence $i$.

\paragraph{Composing sentence representations} We follow a similar method to learn a latent representation of an article. Given the embedding $s_{i}$ of each sentence in the article, we learn a hidden representation of the given sentence $h_{i}$ centered around $s_{i}$ by embedding the list of sentences through a bi-directional GRU as described by \cite{bahdanau2014neural}. Once again, since not all sentences contribute equally to the representation of the article, we introduce a sentence level attention mechanism which attempts to extract relevant sentences that contribute to the meaning of the article. Specifically we learn the weights of a sentence level attention matrix $\bm{W}_{s}$ as $ \alpha_{s}\propto\exp({\bm{W}_{s}h_{s}+b_{s}}),
\bm{z}_{content} = \sum_{s}\alpha_{s}h_{s}$, where $\bm{z}_{content}$ is the latent representation of the article. In this case we let the hidden representation of the sentence be a stochastic representation similar to the work by \cite{miao2016neural} and use the Gaussian re-parameterization trick to enable training via end-to-end gradient based methods \footnote{Using deterministic sentence representations is a special case.}. Such techniques have been shown to be useful in modeling ambiguity and also generalize well to small training datasets \cite{miao2016neural}.

\subsection{Prior} The prior models $p(\bm{h}|\bm{X})$ in Equation \ref{eq:lvb}.  Note that our proposed framework is general and can be used to incorporate a variety of priors. Here, we assume the prior is drawn from a Gaussian distribution with diagonal co-variances. The KL Divergence term in Equation \ref{eq:lvb} can thus be analytically computed. In particular, the KL Divergence between two $K$ dimensional Gaussian distributions $\bm{A, B}$ with means $\mu_{A}, \mu_{B}$ and diagonal co-variances $\kappa_{A}, \kappa_{B}$ is:
\begin{multline}
    D_{KL}(\bm{A}, \bm{B}) = -\frac{1}{2}\sum_{j=1}^{j=K}(1 + \log \frac{\kappa_{Aj}}{\kappa_{Bj}} \\
      - \frac{\kappa_{Aj}}{\kappa_{Bj}} - (\mu_{Aj}-\mu_{Bj})^{2}/\kappa_{Bj})
\end{multline}

\paragraph{Parameter Estimation} Having described precisely, the models for each of the components in Equation \ref{eq:lvb}, we can reformulate the maximization of the variational lower bound to the following loss function on the set of all learn-able model parameters $\theta$:
$\mathcal{J(\theta)}$ as follows:
\begin{equation}
    \mathcal{J}(\theta)=\text{NLL}(\bm{y}|\bm{X})+\lambda \text{D}_{KL}(\bm{q(h)}||\bm{p(h|X)}),
\end{equation} where NLL is the negative log likelihood loss computed between the predicted label and the true label, and $\lambda$ is a hyper-parameter that controls the amount of regularization offered by the KL Divergence term. We use \textsc{AdaDelta} to minimize this loss function. 
\section{Experiments}
We evaluate our model against several competitive baselines which model only a single view to place our model in context:
\begin{enumerate}[noitemsep]
    \item \textbf{Chance Baseline} We consider a simple baseline that returns a draw from the label distribution as the prediction. 
    \item \textbf{Logistic Regression LR (Title)} We consider a bag of words classifier using Logistic Regression that can capture linear relationships in the feature space and use the words of the title as the feature set. 
    \item \textbf{CNN (Title)} We consider a convolutional net classifier based on exactly the same architecture as \cite{kim2014convolutional} which uses the title of the news article. Convolutional Nets have been shown to be extremely effective at classifying short pieces of text and can capture non-linearities in the feature space\cite{kim2014convolutional}.
    \item \textbf{FNN (Network)} We also consider a simple fully-connected feed forward neural network using only the network features to characterize the predictive power of the network alone. 
    \item \textbf{HDAM Model (Content)} We use the state of the art hierarchical document attention model proposed by \cite{yang2016hierarchical} that models the content of the article using both word and sentence level attention mechanisms.
\end{enumerate}
We consider three different flavors of our proposed model which differ in the subset of modalities used (a) \textbf{Title and Network} (b) \textbf{Title and Content}, and (c) \textbf{Full model}: Title, Network, and Content. We train all of our models and the baselines on the training data set choosing all hyper-parameter using the validation set. We report the performance of all models on the held-out test set.

\paragraph{Experimental Settings} We set the embedding latent dimension captured by each view to be $128$ including the final latent representation obtained by fusing multiple modalities. In case of the CNN's, we consider three convolutional filters of window sizes $3,4,5$ each yielding a $100$ dimensional feature map followed by max-over time pooling which is then passed through a fully connected layer to yield the output. In all the neural models, we used AdaDelta with an initial learning rate of $1.0$ to learn the parameters of the model via back-propagation. 
\begin{table}[htb!]
\small
\centering
	\begin{tabular}{l|p{1.75cm}|l|l|l}
	\textbf{Model} & \textbf{Views} & \textbf{P} & \textbf{R} & \textbf{F1} \\
	\hline
	 \textsc{Chance} & -- & 34.53 & 34.59 & 34.53 \\
	 \hline
	 \textsc{LR} & Title & 59.53 & 59.42 & 59.12 \\
	 \textsc{CNN} & Title &  59.26 & 59.40 & 59.24 \\
	 \textsc{FNN} & Network & 68.28 & 56.54 & 55.10 \\

	 \textsc{HDAM} & Content & 69.85 & 68.72 & 68.92 \\
	 \hline
	 \textsc{MVDAM} & Title, Network & \textbf{69.87} & \textbf{69.71} & \textbf{69.66} \\
	 \textsc{MVDAM} & Title, Content & \textbf{70.84} & \textbf{70.19} & \textbf{69.54} \\
	 \textsc{MVDAM} & Title, Network, Content & \textbf{80.10} & \textbf{79.56} & \textbf{79.67} \\
	\hline
	\end{tabular}
	\caption{Precision, Recall, and F1 scores of our model MVDAM on the test set compared with several baselines. All flavors of our model significantly outperform baselines and yield state of the art performance.}
	\label{tab:f1}
\end{table}

\subsection{Results and Analysis}
\paragraph{Quantitative Results} Table \ref{tab:f1} shows the results of the evaluation. First note that the logistic regression classifier and the CNN model using the Title outperforms the \textsc{Chance} classifier significantly (F1: $\bm{59.12, 59.24}$ vs $\bm{34.53}$). Second, only modeling the network structure yields a F1 of $\bm{55.10}$ but still significantly better than the chance baseline. This confirms our intuition that modeling the network structure can be useful in prediction of ideology. Third, note that modeling the content (HDAM) significantly outperforms all previous baselines (F1:$\bm{68.92}$). This suggests that content cues can be very strong indicators of ideology. Finally, all flavors of our model outperform the baselines. Specifically, observe that incorporating the network cues outperforms all uni-modal models that only model either the title, the network, or the content. It is also worth noting that without the network, only the title and the content show only a small improvement over the best performing baseline ($\bm{69.54}$ vs $\bm{68.92}$) suggesting that the network yields distinctive cues from both the title, and the content. Finally, the best performing model effectively uses all three modalities to yield a F1 score of $\bm{79.67}$ outperforming the state of the art baseline by $10$ percentage points. Altogether our results suggest the superiority of our model over competitive baselines.  In order to obtain deeper insights into our model, we also perform a qualitative analysis of our model's predictions.

\begin{figure*}[t!]
\begin{subfigure}{\textwidth}
\centering
	\includegraphics[scale=0.6]{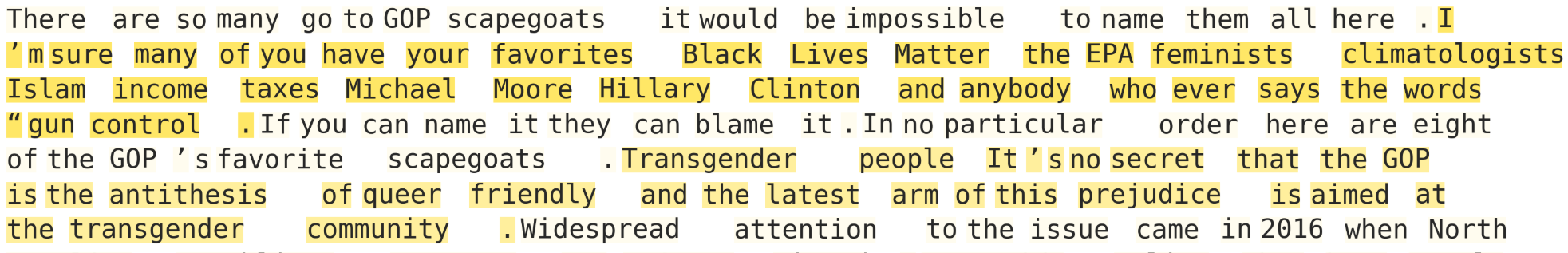}
	\caption{Sample attention on sentences for a Left aligned article.}
	\label{fig:left_wing}
\end{subfigure}
\begin{subfigure}{\textwidth}
\centering
	\includegraphics[width=0.8\textwidth]{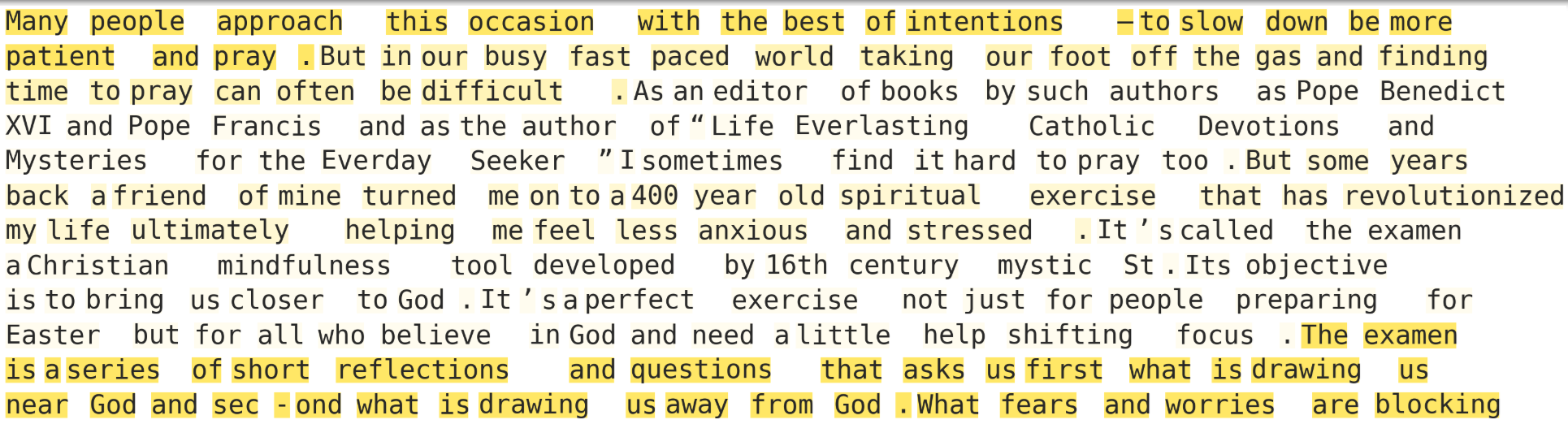}
	\caption{Sample attention on sentences for a Right aligned article.}
	\label{fig:right_wing}
\end{subfigure}
\caption{Visualization of attention on different sentences on two sample articles from the Left and Right aligned sources respectively. Note the different focus based in the ideology reflected by the highlighted words.}
\label{fig:sent_attention}
\end{figure*}

\begin{table*}[htb!]
\small
\centering
\begin{tabular}{p{10cm}|l|l}
\textbf{Article Title} & \textbf{Source Label} & \textbf{Predicted Label} \\ \midrule 
Juan Williams Makes the 'Case for Oprah'  & Right & Left \\
Tourist dies hiking in Australia Outback heat  & Right & Left \\
Back From China, UCLA Basketball Players Plagued by Father & Right & Left \\
Democrat Ralph Northam Elected Governor of Virginia & Right & Left \\
South Africa blighted by racially charged farm murders & Right & Left \\
Lawsuit: Stripper punched man, knocked out his front tooth & Left & Right \\
Here’s How to Keep Fake News Off Twitter & Left & Right \\
We Are All Just Overclocked Chimpanzee & Left & Right \\ 
Curious Arctic Fox Pups Destroy Hidden Camera In The Most Adorable Way & Left & Right \\
I am American, Jewish, and banned from Israel for my activism & Left & Right  \\

\end{tabular}
\caption{Few failure cases of our model illustrating what our model finds challenging. Articles  with ``click-baity'' titles are not necessarily very discriminative of the ideology. Similarly, articles that are \emph{non-political} and related to global events or entertainment are quite challenging.}
\label{tab:slang_errors}
\end{table*}

\begin{table*}[htb!]
\small
	\begin{subtable}[h]{0.33\textwidth}
		\centering
		\begin{tabular}{|l|l|}
		\hline
		\textbf{Rank} & \textbf{Source}\\
		\hline
		1  & CNN\\
		2  & BuzzFeed\\
		3  & SF Chronicle\\
		4  & CBS News\\
		5  & BoingBoing\\
		6  & Mother Jones\\
		7  & Think Progress\\
		8  & The Atlantic\\
		9  & The Washington Post\\
	    10 & Rolling Stone\\
	    \hline
		\end{tabular}
		\caption{Left aligned}
		\label{tab:left_leaning}
	\end{subtable}
	\begin{subtable}[h]{0.33\textwidth}
		\centering
		\begin{tabular}{|l|l|}
		\hline
		\textbf{Rank} & \textbf{Source}\\
		\hline
		1  & NPR\\
		2  & Reuters\\
		3  & USA Today\\
		4  & BBC\\
		5  & CNBC\\
		6  & Chicago Tribune\\
		7  & Business Insider\\
		8  & Forbes\\
		9  & APR\\
	    10  & The Wall Street Journal\\
	    \hline
		\end{tabular}
		\caption{Centre aligned}
		\label{tab:center_leaning}
	\end{subtable}
	\begin{subtable}[h]{0.33\textwidth}
		\centering
		\begin{tabular}{|l|l|}
		\hline
		\textbf{Rank} & \textbf{Source}\\
		\hline
		1  & Breitbart \\
		2  & Infowars \\
		3  & Blaze\\
		4  & Fox News\\
		5  & KSL\\
		6  & Townhall\\
		7  & CBN\\
		8  & ConservativeHQ\\
		9  & NewsMax\\
	    10  & DailyWire\\
	    \hline
		\end{tabular}
		\caption{Right aligned}
		\label{tab:right_leaning}
	\end{subtable}
	\caption{A Top 10 ranking of Ideological sources as obtained by our model which correlate moderately with external surveys.}
	\label{tab:all_leaning}
\end{table*}

\paragraph{Visualizing Attention Scores} Figure \ref{fig:sent_attention} shows a visualization of sentences based on their attention scores. Note that for a left leaning article (see Figure \ref{fig:left_wing}), the model focuses on sentences involving \texttt{gun-control, feminists, and transgender}. In contrast, a visualization of sentence attention scores for an article which the model predicted as ``right-leaning'' ((see Figure \ref{fig:right_wing})) reveals a focus on words like \texttt{god, religion} etc. These observations qualitatively suggest that the model is able to effectively pick up on content cues present in the article. By examining the distribution over the sentence indices corresponding to the maximum attention scores, we noted that only in about half the instances, the model focuses its greatest attention on the beginning of the article suggesting that the ability to selectively focus on sentences in the news article contributes to the superior performance.

\paragraph{Challenging Cases} In Table 2, we highlight some of the challenges of our model. In particular, our model finds it quite challenging to identify the political ideology of the source for articles that are non-political and related to global events, or entertainment. Examples include instances like \texttt{Tourist dies hiking in Australia Outback heat} or \texttt{Juan Williams makes the 'case for Oprah'}. We also note that articles with ``click-baity'' titles like \texttt{We are all Just Overclocked Chimpanzees} are not necessarily discriminative of the underlying ideology. In summary, while our proposed model significantly advances the state of art, it also suggests scope for further improvement especially in identifying political ideologies of articles in topics like Entertainment or Sports. For example, prior research suggests that engagement in particular sports is correlated with the political leanings \cite{hoberman1977sport} which suggest that improved models might need to capture deeper linguistic and contextual cues. 

\paragraph{Ideological Proportions of News Sources} Finally, we compute the expected proportion of an ideology in a given source based on the probability estimates output by our model for the various articles. While one might expect that the expected degree of ``left-ness'' (or ``right-ness'') for a given source can easily be computed by taking a simple mean of the prediction probability for the given ideology over all articles belonging to the source, such an approach can be in-accurate because the probability estimates output by the model are not necessarily calibrated and therefore cannot be interpreted as a confidence value.  We therefore use isotonic regression to calibrate the probability scores output by the model. Having calibrated the probability scores, we now compute the degree to which a particular news source leans toward an ideology by simply computing the mean output score over all articles corresponding to the source. Table \ref{tab:all_leaning} shows the top $10$ sources ranked according to their proportions for each ideology. We note that sources like CNN, Buzz Feed, SF Chronicle are considered more left-leaning than the Washington Post. Similarly, we note that NPR and Reuters are considered to be the most center-aligned while Breitbart, Infowars and Blaze are considered to be most right-aligned by our model. These observations are moderately aligned with survey results that place news sources on the ideology spectrum based on the political beliefs of their consumers \footnote{http://www.journalism.org/2014/10/21/political-polarization-media-habits/pj\_14-10-21\_mediapolarization-08/}.

\section{Conclusion}
\label{sec:conclusion}
We proposed a model to leverage cues from multiple views in the predictive task of detecting political ideology of news articles. We show that incorporating cues from the title, the link structure and the content significantly beats state of the art. Finally, using the predicted probabilities of our model, we draw on methods for probability calibration to rank news sources by their ideological proportions which moderately correlates with existing surveys on the ideological placement of news sources. To conclude, our proposed framework effectively leverages cues from multiple views to yield state of the art interpret-able performance and sets the stage for future work which can easily incorporate other modalities like audio, video and images. 
\section*{Acknowledgments}
We thank the anonymous reviewers for their comments. This research was supported in part by DARPA Grant D18AP00044 funded under the DARPA YFA program. This work was also partially supported by NSF grants DBI-1355990 and IIS-1546113.  The authors are solely responsible for the contents of the paper, and the opinions expressed in this publication do not reflect those of the funding agencies.
\bibliography{acl2018}
\bibliographystyle{acl_natbib}


\end{document}